\documentclass[11pt,a4paper]{article}
\usepackage[hyperref]{acl2020}
\usepackage{times}
\usepackage{latexsym}

\usepackage{microtype}

\aclfinalcopy 

\usepackage{amsmath,graphicx}
\usepackage{algorithmic}
\usepackage{multirow}
\usepackage{url}
\usepackage[caption = false]{subfig}
\usepackage{graphicx}

\title{Multilingual Pre-Trained Transformers and Convolutional NN Classification Models for Technical Domain Identification}

\author{Suman Dowlagar \\
  LTRC\\
  IIIT-Hyderabad\\
  \texttt{suman.dowlagar@} \\
  \texttt{research.iiit.ac.in} \\\And
  Radhika Mamidi \\
  LTRC\\
  IIIT-Hyderabad\\
  \texttt{radhika.mamidi@} \\
  \texttt{iiit.ac.in} \\}

\date{}

\begin{document}
\maketitle
\begin{abstract}
 In this paper, we present a transfer learning system to perform technical domain identification on multilingual text data. We have submitted two runs, one uses the transformer model BERT, and the other uses XLM-ROBERTa with the CNN model for text classification. These models allowed us to identify the domain of the given sentences for the ICON 2020 shared Task, TechDOfication: Technical Domain Identification. Our system ranked the best for the subtasks 1d, 1g for the given TechDOfication dataset.
\end{abstract}

\section{Introduction}

Automated technical domain identification is a
categorization/classification task where the given text is categorized into a set of predefined domains. It is employed in tasks like Machine Translation, Information Retrieval, Question Answering, Summarization, and so on.

In Machine Translation, Summarization, Question Answering, and Information Retrieval, the domain classification model will help leverage the contents of technical documents, select the appropriate domain-dependent resources, and provide personalized processing of the given text.

Technical domain identification comes under text classification or categorization. Text classification is one of the fundamental tasks in the field of NLP. Text classification is the process of identifying the category where the given text belongs. Automated text classification helps to organize unstructured data, which can help us gather insightful information to make future decisions on downstream tasks.

Traditional text classification approaches mainly focus on feature engineering techniques such as bag-of-words and classification algorithms \cite{yang1999evaluation}. Nowadays, the sate-of-the-art results on text classification
are achieved by various NNs such as CNN \cite{kim2014convolutional}, LSTM \cite{hochreiter1997long}, BERT \cite{adhikari2019docbert}, and Text GCN \cite{adhikari2019docbert}. Attention mechanisms \cite{vaswani2017attention}
have been introduced in these models, which
increased the representativeness of the text for better classification. Transformer models such as BERT \cite{devlin2018bert} uses the attention mechanism that learns contextual relations between words or sub-words in a text.  Text GCN \cite{yao2019graph} uses a graph-convolutional network to learn a heterogeneous word document graph on the whole corpus, which helped classify the text. However, of all the deep learning approaches, transformer models provided SOTA results in text classification. 

In this paper, We present two approaches for technical domain identification. One approach uses the pre-trained Multilingual BERT model, and the other uses XLM-ROBERTa with CNN model.

The rest of the paper is structured as follows. Section 2 describes our approach in detail. In Section 3, we provide the
analysis and evaluation of results for our system, and Section 4 concludes our work.

\section{Our Approach}

Here we present two approaches for the TechDOfication task.

\subsection{BERT for TechDOfication}

\begin{figure}[!h]
\centerline{\includegraphics[scale= 0.6]{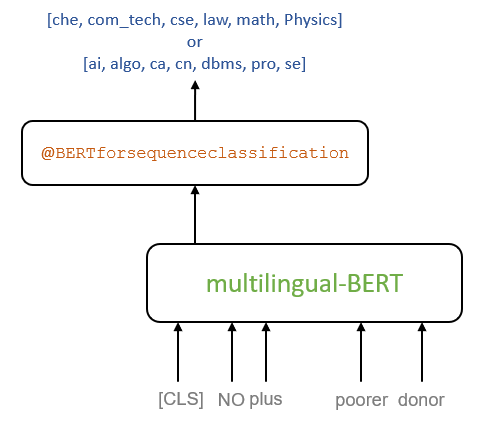}}
\caption{The architecture of the BERT model for sentence classification.}
\label{fig:model}
\end{figure}

In the first approach, we use the pre-trained multilingual BERT model for domain identification of the given text. Bidirectional Encoder Representations from Transformers (BERT) is a transformer encoder stack trained on the large corpora. Like the vanilla transformer model \cite{vaswani2017attention}, BERT takes a sequence of words as input. Each layer applies self-attention, passes its results through a feed-forward network, and then hands it off to the next encoder. The BERT configuration model takes a sequence of words/tokens at a maximum length of 512 and produces an encoded representation of dimensionality 768.

The pre-trained multilingual BERT models have a better word representation as they are trained on a large multilingual Wikipedia and book corpus. As the pre-trained BERT model is trained on generic corpora, we need
to finetune the model for the given domain identification tasks. During finetuning, the pre-trained BERT model parameters are updated.

In this architecture, only the [CLS] (classification) token output provided by
BERT is used. The [CLS] output is the output of the 12th transformer encoder with a dimensionality of 768. It is given as input to a fully connected neural network, and the softmax activation function is applied to the neural network to classify the given sentence.

\subsection{XLM-ROBERTa with CNN for TechDOfication}

\begin{figure}[!h]
\centerline{\includegraphics[scale= 0.6]{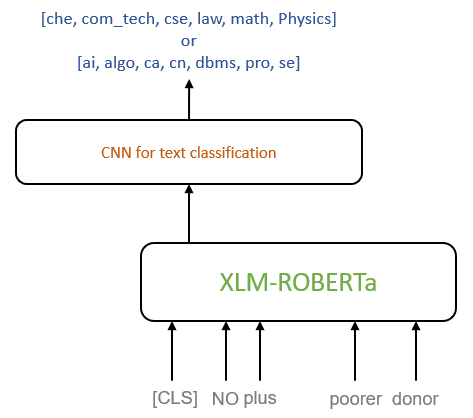}}
\caption{The architecture of the XLM-ROBERTa with CNN for sentence classification.}
\label{fig:model2}
\end{figure}

XLM-ROBERTa \cite{conneau2019unsupervised} is a transformer-based multilingual
masked language model pre-trained on the text in 100
languages, which obtains state-of-the-art performance on cross-lingual classification, sequence labeling, and question answering. XLM-ROBERTa improves upon BERT by adding a few changes to the BERT model such as training on a larger dataset, dynamically masking out tokens compared to the original static masking, and uses a known pre-processing technique (Byte-Pair-Encoding) and a dual-language training mechanism with BERT in order to learn better relations between words in different languages. The given model is trained for the language modeling task, and the output is of dimensionality 768. It is given as input to a CNN \cite{kim2014convolutional} because convolution layers can extract better data representations than Feed Forward layers, which indirectly helps in better domain identification.

\section{Experiment}

This section presents the datasets used, the task description, and two models' performance on technical domain identification. We also include our implementation details and error analysis in the subsequent sections.

\subsection{Dataset}

We used the dataset provided by the organizers of TechDOfication ICON-2020. There are two subtasks, one is coarse-grained, and the other is fine-grained. The coarse-grained TechDOfication
dataset contains sentences about Chemistry, Communication Technology, Computer Science, Law, Math, and Physics domains in different languages such as English, Bengali, Gujarati, Hindi, Malayalam, Marathi, Tamil, and Telugu. Whereas the fine-grained English dataset focuses on the Computer-Science domain with sub-domain labels as Artificial Intelligence, Algorithm, Computer Architecture, Computer Networks, Database Management system, Programming, and Software Engineering. 

\subsection{Implementation}
For the implementation, we used the transformers library provided by HuggingFace\footnote{\url{https://huggingface.co/}}. The HuggingFace contains the pre-trained multilingual BERT, XLM-ROBERTa, and other models suitable for downstream tasks. The pre-trained multilingual BERT model used is \emph{``bert-base-multilingual-cased''} and pre-trained XLM-R model used is \emph{``xlm-roberta-base''}. We programmed the CNN architecture as given in the paper \cite{kim2014convolutional}. We used the PyTorch library that supports GPU processing for implementing deep neural nets. The BERT models were run on the Google Colab and Kaggle GPU notebooks. We trained our classifier with a batch size of 128 for 10 to 30 epochs based
on our experiments. The dropout is set to 0.1, and the Adam optimizer is used with a learning rate of 2e-5. We used the hugging face transformers pre-trained BERT tokenizer for tokenization. We used the BertForSequenceClassification module provided by the HuggingFace library during finetuning
and sequence classification for the multilingual-BERT based approach.

\subsection{Baseline models}

Here, we compared the BERT model with other machine learning algorithms.

\paragraph{SVM with TF\_IDF text representation}
We chose Support Vector Machines (SVM) with TF\_IDF text representation for technical domain identification. SVM classifier and TF\_IDF vector representation is obtained from the scikit-learn library \cite{scikit-learn}.

\paragraph{CNN:} Convolutional Neural Network \cite{kim2014convolutional}. We explored CNN-non-static, which uses pre-trained word embeddings.

\subsection{Results}

The results are tabulated in Table \ref{tab:res}. We evaluated the performance of the method using
macro F1. The multilingual-BERT model performed well when compared to the other SVM with
TF-IDF and CNN models. Given all the languages, we
have observed an increase of 7 to 25\% in classification metrics for BERT 
compared to the baseline SVM classifier, it showed a 2 to 5\% increase in classification metrics compared to the CNN classifier on the validation data. On the test data, multilingual BERT showed better performance in subtasks 1a, 1b, 1c, 1h and 2a whereas XLM-ROBERTa with CNN showed better performance in the subtasks 1d, 1e, 1f, 1g. This increase in classification metrics is due to the transformer model's and convolutional NN's capability, which learned better text representations from the generic data than other models.
\begin{table*}[!t]
\centering
\begin{tabular}{|l|llll|ll|}
\hline
    & \multicolumn{6}{c|}{\textbf{Classifier Models}}                                                                            \\
\hline
           \textbf{Dataset}         & \multicolumn{4}{c|}{\textbf{Validation}}           & \multicolumn{2}{c|}{\textbf{Test}} \\
                    \hline
                    & \textbf{SVM} & \textbf{CNN} & \textbf{M-Bert} & \textbf{XLM-R+CNN} & \textbf{M-Bert}       & \textbf{XLM-R+CNN}       \\
                    \hline
\textbf{English subtask-1a} & 81.48        & 83.05        & \textbf{88.87}           & 87.09                & \textbf{79.84}        & 73.57                      \\
\textbf{Bengali subtask-1b}    & 66.35        & 85.78        & \textbf{86.81}           & 85.71                & \textbf{80.35}        & 78.17                      \\
\textbf{Gujarati subtask-1c}   & 69.63        & 86.27        & \textbf{87.21}           & 86.89                & \textbf{68.67}        & 66.73                      \\
\textbf{Hindi subtask-1d}      & 58.21        & 81.03        & \textbf{83.40}           & 82.13                & 59.89                 & \textbf{60.44}             \\
\textbf{Malayalam subtask-1e}  & 80.60        & 92.51        & \textbf{94.72}           & 93.40                & 34.47                 & \textbf{34.86}             \\
\textbf{Marathi subtask-1f}    & 73.32        & 86.89        & \textbf{87.42}           & 86.37                & 59.52                 & \textbf{59.89}             \\
\textbf{Tamil subtask-1g}      & 65.95        & 85.75        & \textbf{87.50}           & 86.54                & 49.24                 & \textbf{51.34}             \\
\textbf{Telugu  subtask-1h}     & 71.98        & 88.07        & \textbf{90.28}           & 89.43                & \textbf{67.17}        & 62.26                      \\
\textbf{English subtask-2a} & 70.24        & 72.53        & \textbf{77.36}           & 76.77                & \textbf{78.98}        & 78.07   \\
\hline
\end{tabular}
\caption{macro F1 on validation and test data for all the subtasks}
\label{tab:res}
\end{table*}

\section{Error Analysis}

\begin{figure*}[!t]
\centering
\subfloat[]{\includegraphics[scale=0.5]{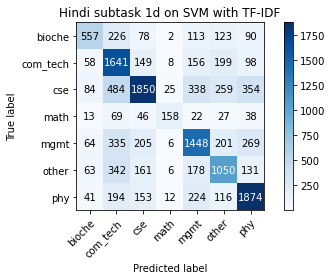}} 
\subfloat[]{\includegraphics[scale=0.5]{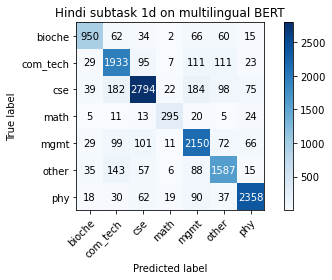}}\\
\subfloat[]{\includegraphics[scale=0.5]{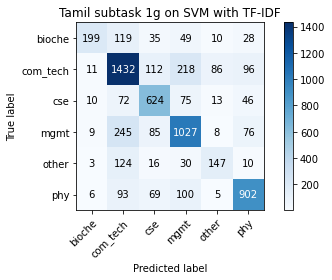}}
\subfloat[]{\includegraphics[scale=0.5]{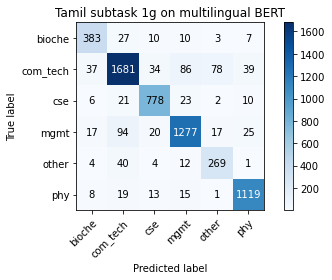}} 
\caption{Confusion matrix on the given validation data for the Hindi and Tamil languages}
\label{fig:cm}
\end{figure*}

The multilingual-BERT model's confusion matrix is compared with the poorly performed model for languages, Hindi, and Tamil languages are shown in Figure \ref{fig:cm}. We chose Hindi and Tamil languages because, here, the difference in performance is more significant. For the Hindi subtask, the SVM classifier confused between ``cse'', ``com\_tech'', and ``mgmt'' labels, whereas the BERT model performed better. For the Tamil subtask, the SVM classifier confused between ``com\_tech'' and ``mgmt'' labels, whereas the BERT model performed better than the other models. This is because both the approaches (pre-trained multilingual-BERT and pre-trained XLM-ROBERTa with CNN) learned better representation of the above data than the other models that helped in technical document identification.

\section{Conclusion and Future work}
We used pre-trained bi-directional encoder representations using multilingual-BERT and XLM-ROBERTa with CNN technical domain identification for English, Bengali, Gujarati, Hindi, Malayalam, Marathi, Tamil, and Telugu languages. We compared the approaches with the baseline methods. Our analysis showed that pre-trained multilingual BERT and XLM-ROBERTa with CNN models and finetuning it for text classification tasks showed an increase in macro F1 score and accuracy metrics compared to baseline approaches.

Some datasets are large, like for the Hindi, Tamil, and Telugu, we can train the BERT and XLM-ROBERTa models from scratch and consider its hidden layer representation, and concatenate this with the representation of the pre-trained model. It might help to classify the datasets even better.

\bibliography{anthology,acl2020}
\bibliographystyle{acl_natbib}

\end{document}